\def\year{2022}\relax
\documentclass[letterpaper]{article} % DO NOT CHANGE THIS
\usepackage{aaai22}  % DO NOT CHANGE THIS
\usepackage{times}  % DO NOT CHANGE THIS
\usepackage{helvet}  % DO NOT CHANGE THIS
\usepackage{courier}  % DO NOT CHANGE THIS
\usepackage[hyphens]{url}  % DO NOT CHANGE THIS
\usepackage{graphicx} % DO NOT CHANGE THIS
\usepackage{natbib}  % DO NOT CHANGE THIS AND DO NOT ADD ANY OPTIONS TO IT
\usepackage{caption} % DO NOT CHANGE THIS AND DO NOT ADD ANY OPTIONS TO IT
\DeclareCaptionStyle{ruled}{labelfont=normalfont,labelsep=colon,strut=off} % DO NOT CHANGE THIS
\usepackage{algorithm}
\usepackage{algorithmic}

\usepackage{newfloat}
\usepackage{listings}
\floatstyle{ruled}
\newfloat{listing}{tb}{lst}{}
\floatname{listing}{Listing}
\usepackage{microtype}
\usepackage{amsmath}
\usepackage{amssymb}
\usepackage{graphicx}
\usepackage{multirow}
\usepackage{longtable}
\usepackage{makecell}
\usepackage{array}

\title{Sequence Level Contrastive Learning for Text Summarization}
\author{Shusheng Xu$^{1}$\thanks{\hspace{0.1cm}~Work done during the first author's internship at \mbox{Microsoft} Research Asia.}~, Xingxing Zhang$^{2}$, Yi Wu$^{1,3}$ and Furu Wei$^2$\\[0.5ex]
	$^1$ IIIS, Tsinghua University, Beijing, China \\
	$^2$ Microsoft Research Asia, Beijing, China \\
	$^3$ Shanghai Qi Zhi Institute, Shanghai China\\
	[0.5ex]
	{\tt xuss20@mails.tsinghua.edu.cn} \\
	{ \tt\{xizhang,fuwei\}@microsoft.com} \\ 
	{\tt jxwuyi@gmail.com}
}
\usepackage{bibentry}
\begin{document}

\maketitle

\begin{abstract}
Contrastive learning models have achieved great success in unsupervised visual representation learning, which maximize the similarities between feature representations of different views of the same image, while minimize the similarities between feature representations of views of different images. In text summarization, the output summary is a shorter form of the input document and they have similar meanings. In this paper, we propose a contrastive learning model for supervised abstractive text summarization, where we view a document, its gold summary and its model generated summaries as different views of the same mean representation and maximize the similarities between them during training. We improve over a strong sequence-to-sequence text generation model (i.e., BART) on three different summarization datasets. Human evaluation also shows that our model achieves better faithfulness ratings compared to its counterpart without contrastive objectives.
\end{abstract}

\section{Introduction}
\label{sec:intro}
Document summarization is the task of rewriting a long document into a shorter form while still preserving its important content, which requires the model to understand the entire document. Many approaches for summarization has been explored in the literature and the most popular ones are \emph{extractive summarization} and \emph{abstractive summarization} \cite{nenkova2011automatic}.
% Extractive summarization, as its name implies, selects (or extracts) the most important sentences in a document as its summary, while abstractive summarization can add new words and phrases or paraphrase the original sentences when generating summaries.
Summaries in their nature are abstractive. The summaries generated by extractive summarization methods are usually long and redundant, which bring bad reading experience. Therefore, we focus on abstractive summarization in this paper. Abstractive summarization is usually modeled as a sequence-to-sequence (Seq2Seq) learning problem \cite{sutskever2014sequence}, where a document is viewed as a sequence of words and its summary another sequence of words \cite{nallapati-etal-2016-abstractive}. 

Although abstractive models have been more and more powerful due to recent introduction of large pre-trained Transformers \cite{liu-lapata-2019-text,raffel2019exploring,dong2019unified,lewis2020bart}, the training paradigm for abstractive models is still not changed, which is to minimize the negative log-likelihood (NLL) between the model predicted word distributions and the gold summary. One great property of the summarization task is that a document and its summary should convey the same meaning, which is not modeled \emph{explicitly} by the NLL  loss.

In computer vision, contrastive learning methods for unsupervised image representation learning advanced the state-of-the-art in object detection and image segmentation \cite{he2020momentum}. The key idea is to minimize distances (or maximize similarities) between feature representations of different views of the same image (positive examples), while to maximize the distances between feature representations of views of different images (negative examples) \cite{he2020momentum,chen2020simple}.
As mentioned earlier, in summarization a document and its summary should convey the same meaning. Therefore, we view a document, its gold summary and its model generated summaries as different views of the same meaning representation and during training, we maximize the similarities between them. To achieve that, we propose SeqCo (as shorthand for \textbf{Seq}uence Level \textbf{Co}ntrastive Learning), which is based on contrastive learning.
In addition to the gold summaries, we also use the dynamically generated summaries from our model during training to increase the diversity of inputs to SeqCo. In text summarization, an abstractive summarization model needs to first encode the document and then generate the summary. The contrastive objective in SeqCo tries to map representations of a document and its summary (or generated summary) to the same vector space, which intuitively helps the generation of summaries.
Specifically, a document may contain distinct (or unnecessary) information from its summary. During training time, the contrastive objective between the document and summary actually encourages the model to encode important (and necessary) information from the document, otherwise the distance between the representations of document and summary will be large (the objective updates model parameters to make it small). Intuitively, the capability of encoding important information from documents would help to generate better summaries. 
% {\bf add sth for contrastive between generated and gold summaries}

In experiments, we find our proposed contrastive learning based model SeqCo consistently improves upon a strong abstractive summarization model based on BART \cite{lewis2020bart} across three different summarization datasets (i.e., CNN/DailyMail \cite{hermann:2015:nips}, New York Times \cite{sandhaus:2008:nyt} and XSum \cite{xsum-emnlp}). 
% \textcolor{red}{We also achieve state-of-the-art results on CNN/DailyMail and New York Times datasets.}
Human evaluation also shows that our model SeqCo achieves better faithfulness ratings compared to its counterpart without contrastive objectives.

\section{Related Work}

The most popular paradigms for summarization are \emph{extractive} and \emph{abstractive} based approaches. 
% As mentioned in Section~\ref{sec:intro}, 
We focus on abstractive summarization. Abstractive summarization may add new words or phrases when generating summaries, which is usually viewed as a sequence to sequence learning problem \cite{nallapati-etal-2016-abstractive,see-etal-2017-get,paulus2018deep,gehrmann:2018:emnlp}. Probably because small and shallow LSTM \cite{hochreiter1997long} based attentive seq2seq models \cite{sutskever2014sequence,bahdanau2014neural} without pre-training are not powerful enough to model documents. Quality of summaries produced by these mdoels are not satisfactory \cite{liu-lapata-2019-text}.
As the recent introduction of large pre-trained transformer models \cite{liu-lapata-2019-text,dong2019unified,zou2020pre,lewis2020bart,zhang2020pegasus,raffel2019exploring}, abstractive models are greatly improved. 
Best results for summarization are achieved by finetuning large models pre-trained with generation (or summarization) tailored objectives on huge amount of unlabeled text ($\ge$160G).  \citet{dong2019unified} pre-train jointly designed Transformer encoder and decoder with language model and masked language model objectives. \citet{zhang2020pegasus} predict gapped sentences from a document removing these sentences and \citet{lewis2020bart} propose sentence permutation and text infilling tasks to pre-train seq2seq transformers. There is also some work on combining extractive and abstractive summarization models \cite{he2020ctrlsum,dou-etal-2021-gsum} or multiple summarization models \cite{liu-etal-2021-refsum}. Unfortunately, pre-training transformers from scratch or combining multiple summarization systems are expensive, while our model can be applied to the light-weighted finetuning stage. 

% \textcolor{cyan}{There are also some works to combine extractive and abstractive methods or combine complementarity of systems' output, \cite{he2020ctrlsum} and \cite{dou-etal-2021-gsum} extract keywords or sentences by an extractive model and then generate the summaries abstractively based on the source documents and extracted tokens. \cite{liu-liu-2021-simcls} and \cite{liu-etal-2021-refsum} train an extra model based on ROUGE \cite{lin-2004-rouge} to rerank and select summary from a set of candidate summaries. These are all combination methods, while we focus on how to train a \emph{single} seq2seq model better.}

Convoluational neural networks pre-trained with contrastive learning methods advance the state-of-the-art in object detection and image segmentation in computer vision \cite{he2020momentum}. The idea is to minimize the distances between feature representations of different views of the same image (positive examples), while to maximize the distances between feature representations of views of different images (negative examples). To discriminate positive examples from negative examples, \citet{he2020momentum} maintain a queue of negative sample representations and utilize momentum updates for encoder of the queue to stabilize these representations. \citet{chen2020simple} use other examples from the same batch as negative examples and as a result, they need a large batch size. These works above suggest that using a large number of negative examples is crucial to obtain good performance, which also increases the complexity for implementation. There is also an interesting line of work without using negative examples. \citet{caron2020unsupervised} employ online clustering to assign codes for two views of the same image and then use representation of one view to predict the cluster codes of the other. During the training of BYOL \cite{grill2020bootstrap}, they only minimized the distance between representations of two views of the same image and they use a momentum encoder for the \emph{target} view to stabilize the training. \citet{chen2020exploring} find that even the momentum encoder can be removed, although there might be a small drop in performance. The contrastive learning method used in our model is most related to BYOL \cite{grill2020bootstrap} in the sense that we do not use negative examples either and we also employ a momentum encoder. In the models above, contrastive learning is applied in the unsupervised pre-training stage, which create different
views of the same image by using effective data argumentation methods. In this paper, we take advantage of the nature
of the summarization task and use the document, gold summary, and generated summary as different views of the same
meaning representation (note that a summary is a shorter form of the original document). To fit sequence-to-sequence learning models for text generation, we handles two sequence of embeddings of discrete words, while the vision models handle two single embeddings of fixed dimensions. In addition, the generated summary are created dynamically during training with a model, which are more diverse than using non-model-based approaches in vision tasks.
% and apply contrastive learning to supervised sequence-to-sequence learning models for text generation. 

In NLP, previously contrastive learning methods are mostly used in pre-training or natural language understanding tasks. For example, {\tt word2vec} \cite{Mikolov2013EfficientEO} learns the word embeddings by distinguishing words in a windows (positive examples) w.r.t. the current word and words randomly sampled (negative examples) using negative sampling. \cite{iter-etal-2020-pretraining} propose a contrastive learning based method for language model pre-training, which predicts the relative distance between sentences using randomly sampled sentences as negative examples. More recently, MatchSum \cite{zhong-etal-2020-extractive} formulates extractive summarization as a semantic text matching problem using contrastive learning. \citet{wu-etal-2020-unsupervised} measures the summary qualities without reference summaries by contrasting the document with the summaries using a ranking model. GSum \cite{dou-etal-2021-gsum} takes different kinds of external guidance as additional input to the document and advances summarization
performance significantly. SimCLS \cite{liu-liu-2021-simcls} proposes a contrastive based
framework for abstractive summarization, which trains a model to rerank the candidate summaries of an abstractive model. We add constrastive learning to the training of an abstractive model by enforcing similarities between document, summary and generated summary, which does not need negative examples. 
\section{Model}
\label{sec:model}
In this section, we describe our contrastive learning model {\bf SeqCo} (as shorthand for {\bf Seq}uence Level {\bf Co}ntrastive Learning) for abstractive text summarization. We first introduce abstractive text summarization models (i.e., Seq2Seq model), on which our model is based. Then we present SeqCo, which adapts contrastive learning to the sequence-to-sequence learning setting.

\subsection{Abstractive Text Summarization}
\label{sec:seq2seq}
% Many text generation tasks in NLP can be modeled as sequence-to-sequence (Seq2Seq) learning problems. For example, 
For text summarization, we can view the document as a long  sequence of tokens\footnote{We use \emph{tokens} instead of \emph{words}, because the sequence might be a sequence of sub-words.} and the summary as a short sequence of tokens. Let $X = (x_0=\text{\tt <s>}, x_1, x_2, \dots, x_{|X|}=\text{\tt </s>})$ denote a document (i.e., the long sequence  of tokens) and $Y = (y_0=\text{\tt <s>}, y_1, y_2, \dots, y_{|Y|} =\text{\tt </s>} )$ its summary (i.e., the short sequence of tokens), where $\text{\tt <s>}$ and $\text{\tt </s>}$ are begin and end of sequence tokens. We predict $Y$ one token at a time given $X$. We adopt the Transformer model \cite{vaswani:2017:nips}, which is composed of an encoder Transformer and a decoder Transformer. Specifically, the encoder Transformer  maps $X$ into a sequence of hidden states $\mathbf{E} = (\mathbf{e}_0, \mathbf{e}_1, \dots, \mathbf{e}_{|X|})$. 
\begin{equation}
    \label{eq:enc}
    \mathbf{E}  = \mathrm{Trans}^{\text{E}}(X)
\end{equation}
Supposing that the first $t-1$ tokens $y_{1:t-1}$ have been generated and we are generating $y_t$.  The decoder Transformer computes the current hidden state $\mathbf{o}_t$ by \emph{self} attending to the encoder hidden states $\mathbf{E}$ and proceeding tokens $y_{0:t-1}$.
\begin{equation}
\label{eq:dec}
\mathbf{o}_t = \mathrm{Trans}^{\text{D}}(y_{0:t-1}, \mathbf{E})
\end{equation}
Note that during training, we can obtain $\mathbf{O} = (\mathbf{o}_1, ..., \mathbf{o}_{|Y|)}$ in parallel.
\begin{equation}
\label{eq:dec_all}
\mathbf{O} = \mathrm{Trans}^{\text{D}}(Y, \mathbf{E})
\end{equation}
The probability of $y_t$ can be estimated using a linear projection and a \emph{softmax} function
\begin{equation}
p(y_t | y_{0:t-1}, X) = \text{softmax}( \mathbf{W}^o \, \mathbf{o}_t)
\end{equation}

\begin{equation}
    \label{eq:nll_loss}
    \mathcal{L}^{\text{NLL}} =  - \frac{1}{|Y|} \sum_{t=1}^{|Y|} \log p(y_t |y_{0:t-1}, X)
\end{equation}
% where $\mathcal{D}$ is the training corpus.

\subsection{SeqCo: Sequence Level Contrastive Learning for Text Summarization}
\label{sec:contrastive}
% In text generation tasks, we use the input sequence $X$ to infer the output sequence $Y$ and they usually share similar meanings. For example, 
\begin{figure}[t]
    \centering
    \includegraphics[width=0.65\linewidth]{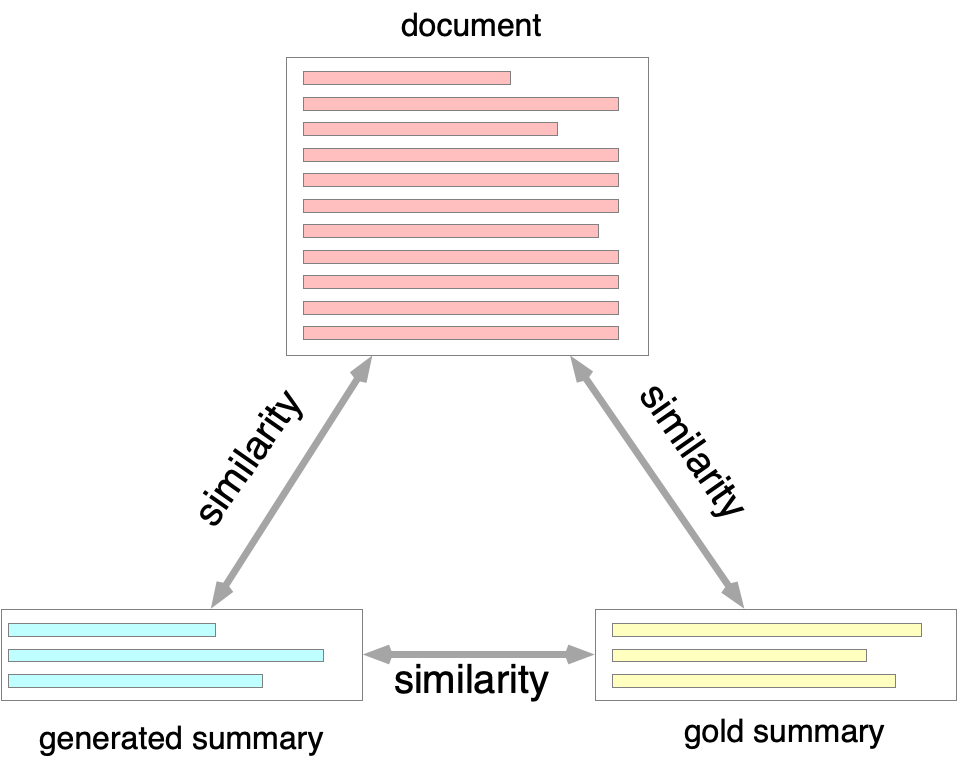}
    \caption{We enforce the similarities between the document, gold summary and model generated summary.}
    \label{fig:pull_all}
\end{figure}

In text summarization, the summary $Y$ is a shorter form of the input document $X$ and they should convey the same meaning. 
% In text summarization, the summary $Y$ convey the same meaning as the input document $X$, and $Y$ is a shorter form of $X$. 
Therefore, $X$ and $Y$ should be close in the semantic space at least after certain types of transformations. However, a Seq2Seq model is trained using the negative log-likelihood loss (see Equation (\ref{eq:nll_loss})) and there is no explicit modeling for the similarity between $X$ and $Y$. Further, during the training phase, given $X$  as input, the model can also generate output sequences from its distribution by either beam search or sampling. Let $\hat{Y}$ denote one sample the model generated from $X$. Intuitively, $\hat{Y}$ should also be similar to both $X$ and $Y$. As shown in figure~\ref{fig:pull_all}, we enforce the similarities between $X$, $Y$ and $\hat{Y}$ during model training. To do this, we propose \mbox{SeqCo}, which is a contrastive learning based model for text summarization.

Contrastive learning methods are proposed in the context of self-supervised learning for image representations \cite{wu2018unsupervised,he2020momentum,caron2020unsupervised,grill2020bootstrap,chen2020exploring}. The training objective tries to make representations of different views of the same image  closer (positive examples) while representations of views of different images apart from each other (negative examples). 
% To leverage negative examples, a queue or memory bank is maintained \cite{wu2018unsupervised,he2020momentum}, which introduces additional memory (computional) cost and strategies for selecting negative examples. 
Inspired by \citet{grill2020bootstrap} and \citet{chen2020exploring}, we propose a model that does not need negative examples. In the following, we first define similarity measures between sequences and then we present how to equip the similarity measures into our training objective.

\begin{figure}[t]
    \centering
    \includegraphics[width=0.8\linewidth]{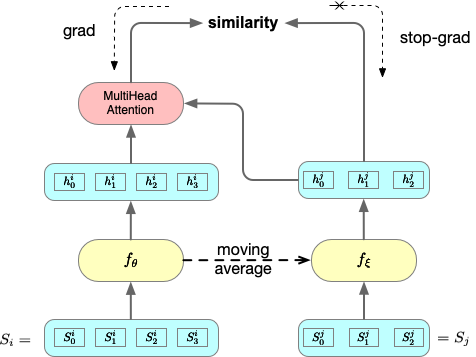}
    \caption{The contrastive objective. $S_i$ and $S_j$ are two sequences to contrast, $f_\theta$ and $f_\xi$ have the same architecture, $\theta$ in $f_\theta$ is updated by gradient decent while $\xi$ in $f_\xi$ is the moving average of $\theta$.}
    \label{fig:sim}
\end{figure}

\paragraph{Sequence Representation} Suppose that we have two sequences $S_i = (w_0^i, w_1^i, w_2^i, ..., w_{|S_i|}^i)$ and $S_j = (w_0^j, w_1^j, w_2^j, ..., w_{|S_j|}^j)$. $S_i$ and $S_j$ are two sequences, which we will maximize their similarity in Eq.~\ref{eq:final_loss}. For example, $S_i$ and $S_j$ can be a document X and its gold summary Y, or document and generated summary, or gold summary and generated summary, just like Fig.~\ref{fig:sim}. Before going to the similarity computation, we first convert them into sequences of hidden representations. We designed two mapping functions here. The first one ($f_\theta^{\text{E}}$) is unconditional, which \emph{reuses} the encoder of our Seq2Seq model (see Section \ref{sec:seq2seq}):
\begin{equation}
\label{eq:fenc}
f_\theta^{\text{E}} (S_i) = g ( \mathrm{Trans}^{\text{E}}(S_i) )
\end{equation}
where $\mathrm{Trans}^{\text{E}}(\cdot)$ is the Transformer encoder described in Equation (\ref{eq:enc}) and $g(\cdot)$ is a feed-forward network that is used to give more freedom for encoding $S_i$. Here we use $\theta$ to denote the parameters in $f_\theta^{\text{E}} (\cdot)$. 
 
The second mapping function ($f_\theta^{\text{D}}$) is conditional, which takes of the input sequence into account.\footnote{Note that in $f_\theta^{\text{D}}$ we only consider that $S_i$ and $S_i$ as the gold summary and the generated summary} Let $X$ denote the input sequence and $S_i$ is its \emph{gold} output sequence or a sequence generated by the Seq2Seq model. In this mapping function, we employ both the encoder and the decoder of the Seq2Seq model (see Section \ref{sec:seq2seq} for details):
\begin{equation}
\label{eq:fdec}
f_\theta^{\text{D}} (S_i) = g( \mathrm{Trans}^{\text{D}}(S_i, \mathrm{Trans}^{\text{E}}(X) ) )
\end{equation}
where $\mathrm{Trans}^{\text{E}}(\cdot)$ and $\mathrm{Trans}^{\text{D}}(\cdot)$ are the Transformer encoder and decoder described in Equation (\ref{eq:enc}) and (\ref{eq:dec_all}). As mentioned earlier, $g(\cdot)$ is a feed-forward network to give more freedom for encoding $S_i$. %\xss{we also use $\theta$ to denote the parameters in $f_\theta^{\text{D}} (\cdot)$} . 
In $f_\theta^{\text{D}}(\cdot)$, we intend to use $X$ as additional input to encode $S_i$ more accurately in vector space. During contrastive training, using $f_\theta^{\text{D}}(\cdot)$ can force the objective to optimize both the encoder and the decoder of the summarization model. 

\paragraph{Sequence Similarity}
After defining the mapping functions, we are ready to compute sequence similarities. Without losing generality, let $f_\theta$ denote the mapping function, where $\theta$ is the parameter of the function. 
Note that $f_\theta $ can be either $f_\theta^{\text{E}}$ or $f_\theta^{\text{D}}$ (see Eq. (\ref{eq:fenc}) and (\ref{eq:fdec}) for details). We additionally employ another mapping function $f_{\xi}$, which has the same architecture as $f_{\theta}$, but with parameter $\xi$. 
% We will explain the reason why we use different parameters later. 
We obtain the representations of $S_i$ and $S_j$ by applying $f_{\theta}$ and $f_{\xi}$ to them:
\begin{equation}
\begin{split}
\mathbf{H}^i &= (\mathbf{h}_0^i, \mathbf{h}_1^i, \dots, \mathbf{h}_{|S_i|}^i) = f_{\theta}(S_i) \\
\mathbf{H}^j &= (\mathbf{h}_0^j, \mathbf{h}_1^j, \dots, \mathbf{h}_{|S_j|}^j) = f_{\xi}(S_j)
\end{split}
\end{equation}
To fully utilize the word-to-word interactions between the two sequences $S_i$ and $S_j$, we apply a cross attention between $\mathbf{H}^i$ and $\mathbf{H}^j$:
\begin{equation}
\label{eq:multi-head}
\widetilde{ \mathbf{H}^i } = \mathrm{MultiHeadAttn}(\mathbf{H}^j, \mathbf{H}^i, \mathbf{H}^i)
\end{equation}
where $ \mathrm{MultiHeadAttn}(\cdot, \cdot, \cdot)$ is the multi-head \mbox{attention} module \cite{vaswani:2017:nips} and $\mathbf{H}^j$, $\mathbf{H}^i$ and $\mathbf{H}^i$ are the query, key and value matrices, respectively. Note that the resulting $\widetilde{ \mathbf{H}^i } $ and $\mathbf{H}^j$ have the same size.
The similarity between $S_i$ and $S_j$ is the averaged cosine similarities of all vectors with the same index:
\begin{equation}
\label{eq:sim}
sim(S_i, S_j) = \frac{1}{|S_j| + 1} \sum_{k=0}^{|S_j|} \cos(  \widetilde{  \mathbf{h}_k^{i} } ,  \mathbf{h}_k^j) 
\end{equation}
We adopt multi-head attention (MHA) for similarity computation  for two reasons. 1) The sequences (esp. documents) are long and MHA takes all pairs of tokens across two sequences into account, which is intuitively more powerful than {\tt [CLS]} pooling based methods (will introduce below). 2) The two sequences we compare may have different lengths (e.g., a document v.s. a summary). MHA can convert the hidden states of one sequence to the same length as the hidden states of another sequence (see Equation \ref{eq:multi-head}), which are easier to use for the similarity computation.

Note that we can also define a simpler similarity function using the {\tt [CLS]} pooling as in BERT \cite{devlin2018bert}:
\begin{equation}
\label{eq:sim_cls}
sim(S_i, S_j)=\cos(q(\mathbf{h}_0^i), \mathbf{h}_0^j)
\end{equation}
where $q$ is a feed-forword network to project $\mathbf{h}_0^i$ following \citet{grill2020bootstrap}. We obtained worse results using the similarity measure above (see Section \ref{sec:results} for details) and the measure also sometimes leads to numerical errors during training. 

\paragraph{Training} To make $S_i$ and $S_j$ closer, we can minimize the following loss:
\begin{equation}
 \mathcal{L}_{\theta, \xi}(S_i, S_j) = 1 - sim(S_i, S_j)
\end{equation}
As mentioned earlier, $f_{\theta}$ (the encoding function for $S_i$) and $f_{\xi}$ (the encoding function for $S_j$) use different set of parameters (i.e., $\theta$ and $\xi$). If we update the parameters in both $f_{\theta}$ and $f_{\xi}$ simultaneously, the optimization maybe too easy, which may lead to collapsed solutions \cite{grill2020bootstrap}. So we use $f_{\xi}$ to produce regression \emph{targets} for $f_{\theta}$. Specifically, we do not update the parameters in $f_{\xi}$ during the optimization of the loss above and $\xi$ is a moving average of  $\theta$:
 \begin{equation}
 \label{eq:moving_average}
 \xi = \tau \xi + (1 - \tau) \theta
 \end{equation}
 where $\tau \in [0, 1]$ is a hyper-parameter to control the extend of retaining $\xi$. This contrastive objective is demonstrated in figure~\ref{fig:sim}.
 % {\bf maybe we say the range for $\tau$ \xss{We have said  $\tau \in [0, 1]$, need a little more detail?}}
 Note that $ \mathcal{L}_{\theta, \xi}(S_i, S_j) $ is not symmetric and we make the loss symmetric as follows: % {\bf need better justification}
 \begin{equation}
 \label{eq:sim_loss}
 \mathcal{L}_{\text{sim}} ( S_i, S_j )  = \mathcal{L}_{\theta, \xi}(S_i, S_j) + \mathcal{L}_{\theta, \xi}(S_j, S_i)
 \end{equation}
 Hence, $\theta$ in $f_{\theta}$ will have more chances to be updated.
% \xss{Since that we $S_i$ and $S_j$ should be similar, $f_\theta$ and $f_\xi$ is not exactly the same, this symmetric form is adopted to increase the training data.}
As mentioned earlier, the encoding function $f_{\theta}$ can be either $f_\theta^{\text{E}}$ or $f_\theta^{\text{D}}$. We use $\mathcal{L}_{\text{sim}}^{\text{E}} $ to denote the loss function using $f_\theta^{\text{E}}$ and $\mathcal{L}_{\text{sim}}^{\text{D}} $  to denote the loss function using $f_\theta^{\text{D}}$.

To enforce the similarities between the document $X$, its gold summary $Y$ and one of the model generated summary $\hat{Y}$, we employ the following loss function as our final training loss\footnote{We can also use multiple generated summaries in training, we refrained to do so for efficiency reasons.}:
\begin{equation}
\label{eq:final_loss}
\begin{split}
\mathcal{L} =&  \mathcal{L}^{\text{NLL}} + \lambda_{x-y} \mathcal{L}_{\text{sim}}^{\text{E}}(X, Y) + \lambda_{x-\hat y} \mathcal{L}_{\text{sim}}^{\text{E}}(X, \hat{Y}) \\
&+ \lambda_{y-\hat y} \mathcal{L}_{\text{sim}}^{\text{E}}(Y, \hat{Y}) + \lambda^{\text{D}}_{y-\hat y} \mathcal{L}_{\text{sim}}^{\text{D}}(Y, \hat{Y})
\end{split}
\end{equation}
This objective contains five terms. $\mathcal{L}^{\text{NLL}}$ is the negative log-likelihood;  $\mathcal{L}_{sim}^D$ is the similarity loss w.r.t. $(Y, \hat{Y})$ with $f_\theta^{\text{D}}$; $\mathcal{L}_{sim}^{\text{E}}$ terms are the similarity losses with $f_\theta^{\text{E}}$ w.r.t. $(X, Y)$, $(X, \hat{Y})$ and $(Y, \hat{Y})$. $\lambda_{x-y}$, $\lambda_{x-\hat y}$, $\lambda_{y-\hat y}$ and $\lambda^{\text{D}}_{y-\hat y}$ are weight hyper-parameters for the last four terms. 
We completely train the model end-to-end following this loss function and empirically find that using a \emph{single} similarity loss works better than using multiple ones (see Section~\ref{sec:results}), which is also more efficient for training. For example, we can set $\lambda_{x-\hat y} = 1.0$ and $\lambda_{x-y}=\lambda_{y-\hat y}=\lambda^{\text{D}}_{y-\hat y}=0$. When $\hat{Y}$ is adopted, The model iteratively generates $\hat{Y}$ by using the loss to update parameters and generating new $\hat{Y}$. Since $\hat{Y}$ can not be perfect, iteratively generating $\hat{Y}$ makes it
change toward ground-truth summary and make the positive
examples for contrastive learning more accurate and diverse. Since SeqCo is designed for the fine-tuning stage, and the model SeqCo based on (i.e., BART) is pre-trained with a denoising auto-encoding objective, it can naturally generate the sequence with the same meaning as the input even before fine-tuning in a specific dataset.
In addition, enforcing the similarity of $y$ and $\hat{y}$ does not equals optimizing NLL, since the similarity loss is on sequence level while the NLL loss is on token level.

\section{Experiments}

In this section, we assess the preformance of our contrastive learning model on the task of text summarization. We will first introduce the datasets we used. Then we present our implementation details. Finally, we compare our model with multiple previous models.

\subsection{Datasets}
\paragraph{CNNDM} We conduct our experiments on three summarization datasets. The CNN/DailyMail dataset (\mbox{CNNDM}; \citealt{hermann:2015:nips}) contains news articles and their associated highlights (i.e., reference summaries) from the CNN and Daily Mail websites. We follow the standard pre-processing steps in \cite{see-etal-2017-get}\footnote{Available at https://github.com/abisee/cnn-dailymail} and the resulting dataset contains 287,226 articles for training, 13,368 for validation and 11,490 for test. 

\paragraph{NYT} The New York Times dataset (\mbox{NYT}; \citealt{sandhaus:2008:nyt}) is composed of articles published by the New York Times with summaries written by library scientists. Following the pre-processing procedures in \cite{durrett-etal-2016-learning,liu-lapata-2019-text}, we first obtain 110,540 articles with abstractive summaries. The test set is constructed from the 9,706 articles published after January 1, 2007. After removing articles whose summaries are shorter than 50 words, the final test set contains 3,452 articles. The remaining 100,834 articles are filtered and splitted into 38,264 articles for training and 4,000 articles for validation.

\paragraph{XSum} The articles in the XSum dataset \cite{xsum-emnlp} are from the BBC website with accompanying single sentence summaries, which are professionally written. We use the official splits of \cite{xsum-emnlp} (i.e., 204,045 articles for training, 11,332 articles for validation and 11,334 articles for test). 
\\
\\
All datasets are tokenized with the byte-pair encoding of GPT2 \cite{radford2019language}.

\subsection{Implementation Details}
Our model is initialized from $\text{BART}_{\text{Large}}$ \cite{lewis2020bart}. Therefore, the size is identical with $\text{BART}_{\text{Large}}$ \cite{lewis2020bart}. Specifically, the encoder and decoder are all 12-layer transformers with 16 attention heads, hidden size 1,024 and feed-forward filter size 4,096, which amounts to 406M trainable parameters. We also have additional component for contrastive learning. The feedforward network $g$ (see Equation~(\ref{eq:fenc}) and~(\ref{eq:fdec})) for projecting sequence features contains one hidden layer of 4,096 neurons with ReLU activation function. The multi-head attention module (see Equation~(\ref{eq:multi-head})) used to compute cross attention between sequences also has 16 heads. These two components above contribute to an extra 13M trainable parameters.

We optimize the model using Adam with  $\beta_1 = 0.9, \beta_2 = 0.999$. Following \cite{lewis2020bart}, we employ a linear schedule for the learning rate. We firstly warmup the model by increasing the learning rate linearly to a peak learning rate and then decrease the learning rate linearly to zero. The peak learning rate, warmup steps, total number of updates and batch size are tuned on validation sets and are different across datasets, which are $1000$, $20000$, $4e-5$, $128$ on CNNDM, $500$, $5000$, $2e-5$, $64$ on NYT, $500$, and $15000$, $6e-5$, $64$ on XSum. In all datasets, the number of training epochs are between 5 to 10.
During the optimization, parameters $\xi$ in the online encoding function $f_\xi$ (see Equation~(\ref{eq:fenc}) and~(\ref{eq:fdec})) are not updated. Parameters $\xi$ in$f_\xi$  are updated following Equation~(\ref{eq:moving_average}) with $\tau = 0.99$. We employ label smoothing of 0.1 \cite{Szegedy2016RethinkingTI,vaswani:2017:nips}. 
The models for CNNDM are trained on 8 Tesla V100 GPUs, and the models for the other datasets are trained on 4 Tesla V100 GPUs. During decoding, we select minimum generated length and length penalty according to ROUGE scores on the validation set. Following \cite{paulus2018deep}, we also blocked repeated trigrams during beam search. Following \cite{lewis2020bart}, the articles are truncated to 1024 tokens in both training and decoding.

\subsection{Evaluations}

We use ROUGE \cite{lin-2004-rouge} to measure the quality of generated summaries. We reported full-length F1 based ROUGE-1, ROUGE-2 and ROUGE-L scores on CNNDM and XSum datasets. 
Following \cite{durrett-etal-2016-learning}, we use the limited-length recall based ROUGE-1, ROUGE-2 and ROUGE-L on NYT, where generated summaries are truncated to the length of gold summaries. ROUGE scores are computed with the {\tt ROUGE-1.5.5.pl} script\footnote{with -c 95 -r 1000 -n 2 -a -m arguments}.

\subsection{Results}
\label{sec:results}
We present our main results on the CNNDM dataset in Table~\ref{tab:cnndm}. We compare our model against both extractive and abstractive systems. The first block summarizes the results for extractive systems. Lead3 is a baseline which simply takes the leading three sentences in a document as its summary. {\sc BertExt} \cite{liu-lapata-2019-text} employs BERT as encoder and predicts whether a sentence is a summary. MatchSum \cite{zhong-etal-2020-extractive} is the best performing extractive models, which formulates summarization as a semantic text matching problem using contrastive learning. 
The abstractive models are in the second block. PTGen \cite{see-etal-2017-get} is a LSTM-based Seq2Seq model augmented with copy and coverage models. Large pre-trained language models mostly dominate summarization. {\sc BertSumExtAbs} \cite{liu-lapata-2019-text} is an abstractive model with encoder initialized with BERT and decoder randomly initialized. UniLM \cite{dong2019unified} is trained using language modeling and masked language modeling objectives. T5 \cite{raffel2019exploring}, PEGASUS \cite{zhang2020pegasus}, BART \cite{lewis2020bart} and STEP \cite{zou2020pre} pre-train Seq2Seq transformers using different unsupervised text-to-text tasks. PEGASUS \cite{zhang2020pegasus} is trained by predicting gapped sentences (selected by some heuristics) in a document given the document with these sentences masked. Similar to {\sc BertSumExtAbs}, the encoder of STEP is initialized from RoBERTa \cite{liu2019roberta}. BART + R3F \cite{aghajanyan2021better} applies a trust region theory based fine-tuning method to BART. Our model is based on BART and therefore we also re-implement BART (BART$\star$). These models above are single models. We also present the results of recent combination models in the third block. CTRLsum \cite{he2020ctrlsum} and GSum \cite{dou-etal-2021-gsum} combine a keywords extraction model (or an extractive model) with an abstractive model by taking the resulting keywords (or sentences) as additional input. SimCLS\cite{chen2020simple} and Refsum \cite{liu-etal-2021-refsum} train re-ranking models to rank multiple candidate summaries.
% from multiple abstractive models.

% extract keywords or sentences by an extractive model and then generate the summaries abstractively based on the source documents and extracted tokens.  Refsum \cite{liu-etal-2021-refsum} train an extra model based on ROUGE \cite{lin-2004-rouge} to rerank and select summary from a set of candidate summaries. These methods advence the SOTA results by combining several models or reranking system outputs, while we focus on how to train a \emph{single} seq2seq model better.

The fourth block includes results of our model \mbox{SeqCo}. As mentioned in Section \ref{sec:contrastive}, we can do contrastive learning between document and gold summary (i.e., SeqCo ($\lambda_{x-y}$)), document and generated summary (i.e., SeqCo ($\lambda_{x-\hat{y}}$)) as well as gold summary and generated summary (i.e., \mbox{SeqCo ($\lambda_{y-\hat{y}}$)}). Note SeqCo ($\lambda_{*-*}$) means that $\lambda_{*-*} > 0$ and all the other $\lambda$s equal to zero in Equation~(\ref{eq:final_loss})\footnote{We tune $\lambda_{x-y},\lambda_{x-\hat{y}},\lambda_{y-\hat{y}} \in \{0.5, 1.0\}$ on the validation set when $> 0$}. We can see that SeqCo ($\lambda_{x-y}$), SeqCo ($\lambda_{x-\hat{y}}$) and SeqCo ($\lambda_{y-\hat{y}}$) all outperform BART$\star$ significantly ($p<0.05$) measured by the ROUGE script, which demonstrates the effectiveness of our proposed contrastive methods. \mbox{SeqCo ($\lambda_{y-\hat{y}}$)} outperforms all \emph{single} models in comparison (first two blocks) and differences between them are significant w.r.t. the ROUGE script. We also observe that using generated summaries in contrastive learning leads to better performance (i.e., results of \mbox{SeqCo ($\lambda_{x-\hat{y}}$)} and SeqCo ($\lambda_{y-\hat{y}}$) are better), which is not surprising. Generated summaries are created dynamically during training and they might be more diverse than gold summaries. 

\begin{table}[ht]
	\small
	\centering
	\begin{tabular}{l c c c}
		\hline
		Model &  R-1 & R-2 & R-L \\
		\hline 
		\multicolumn{4}{c}{Extractive}  \\
		\hline
		Lead3 & 40.34 & 17.70 &  36.57 \\
		{\sc BertExt} \cite{liu-lapata-2019-text}  & 43.85 &  20.34 & 39.90 \\
		{\sc MatchSum} \cite{zhong-etal-2020-extractive} & \textbf{44.41} & \textbf{20.86} & \textbf{40.55} \\
		\hline
		\multicolumn{4}{c}{Abstractive} \\
		\hline
% 		PTGen \cite{see-etal-2017-get} & 39.53 & 17.28 & 36.38 \\
        PTGen \shortcite{see-etal-2017-get} & 39.53 & 17.28 & 36.38 \\
		{\sc BertSumExtAbs}  \shortcite{liu-lapata-2019-text} & 42.13 & 19.60 & 39.18 \\
		UniLM \cite{dong2019unified} & 43.47 & 20.30 & 40.63 \\
		% TRANSFORMER-S2S \cite{zou2020pre}  & 40.43 & 17.66 & 37.44 \\
		T5 \cite{raffel2019exploring} & 43.52 & \textbf{21.55} & 40.69 \\
		PEGASUS (C4) & 43.90 & 21.20 & 40.76 \\
		PEGASUS (HugeNews) &  44.17 & 21.47 & 41.11 \\
		STEP \cite{zou2020pre} & 44.03 & 21.13 & 41.20 \\
		BART \cite{lewis2020bart} & 44.16 & 21.28 & 40.90 \\
		BART$\star$ \cite{lewis2020bart} & 44.10 & 21.31 & 40.91 \\
		BART + R3F \shortcite{aghajanyan2021better} & \textbf{44.38} & 21.53 & \textbf{41.17} \\
		\hline
		\multicolumn{4}{c}{{Combination Methods}} \\
		\hline
		CTRLsum \cite{he2020ctrlsum} & 45.65 & 22.35 & 42.50 \\
		GSum \cite{dou-etal-2021-gsum} & 45.94 & 22.32 & 42.48 \\
		simCLS \cite{liu-liu-2021-simcls} & \textbf{46.67} & 22.15 & 43.54 \\
		Refsum \cite{liu-etal-2021-refsum} & 46.12 & \textbf{22.46} & \textbf{42.92} \\
		\hline
		\multicolumn{4}{c}{Ours} \\
		\hline
		SeqCo ($\lambda_{x-y}$) & 44.66\dag & 21.57* & 41.38* \\
		SeqCo ($\lambda_{x-\hat{y}}$) & 44.94\dag & \textbf{21.82}\dag & 41.68\dag  \\
		SeqCo ($\lambda_{y-\hat{y}}$) & \textbf{45.02}\dag & 21.80\dag & \textbf{41.75}\dag \\
		\hline
	\end{tabular}
	\caption{Results on the test split of CNNDM using full length F1 based ROUGE-1/2/L. $\star$ means our own re-implementation. SeqCo ($\lambda_{x-y}$), SeqCo ($\lambda_{x-\hat{y}}$) and SeqCo ($\lambda_{y-\hat{y}}$) stand for contrastive learning between document and gold summary, document and generated summary as well as gold and generated summary, respectively. * means outperforms BART$\star$ significantly, \dag\ means outperforms best performing single model ``BART+R3F'' significantly ($p < 0.05$). Models in ``Combination Methods'' employ multiple summarization models.}
	\label{tab:cnndm}
\end{table}

\begin{table}[ht]
	\small
	\centering
	\begin{tabular}{l c c c}
		\hline
		Model &  R-1 & R-2 & R-L \\
		\hline
		BART$\star$ & 45.24 & 22.10 & 42.01 \\
		SeqCo ($\lambda_{x-y}$) & 45.60 & 22.30 & 42.36 \\
		SeqCo ($\lambda_{x-\hat{y}}$) & 45.80 & 22.39 & 42.57  \\
		% BART + $\lambda_b(<s>)$ &  \\
		SeqCo ($\lambda_{y-\hat{y}}$) & \textbf{45.88} & \textbf{22.46} & \textbf{42.66} \\
		SeqCo ($\lambda_{y-\hat{y}}$) w/ {\tt [CLS]} & 45.72 & 22.42 & 42.48 \\
		SeqCo ($\lambda_{x-y}$ + $\lambda_{x-\hat{y}}$) & 45.68 & 22.38 & 42.45 \\
		SeqCo ($\lambda_{x-y}$ + $\lambda_{y-\hat{y}}$)& 45.62 & 22.29 & 42.37 \\
		SeqCo ($\lambda_{x-\hat{y}}$ + $\lambda_{y-\hat{y}}$) & 45.72 & 22.35 & 42.45 \\
		SeqCo ($\lambda_{x-y}$ + $\lambda_{x-\hat{y}}$ + $\lambda_{y-\hat{y}}$) & 45.72 & 22.38 & 42.46 \\
		SeqCo ($\lambda^D_{y-\hat{y}}$) & 45.74 & 22.39 & 41.55 \\
		\hline
	\end{tabular}
	\caption{Results on the validation split of CNNDM using full length F1 based ROUGE-1/2/L. ``w/ [CLS]'' means we replace MHA with [CLS] pooling defined in Eq.~\ref{eq:sim_cls}}.
	\label{tab:dev_cnndm}
\end{table}

It is also possible to employ multiple pairs of text for contrastive learning. Results on validation set with different combinations of text pairs are shown in Table \ref{tab:dev_cnndm}. We obtain worse results with more than one pair of text in contrastive learning. Perhaps because the information learned using different pair of text is a bit redundant. We compared the results on the validation and test sets of the other two datasets and observed similar trends.\footnote{ Detailed numbers are shown in Appendix.} We find best results are achieved by using a single similarity loss on all datasets except for the validation set of XSum, where SeqCo ($\lambda_{x-y}$ + $\lambda_{y-\hat{y}}$)  and SeqCo ($\lambda_{x-y}$ + $\lambda_{x-\hat{y}}$ + $\lambda_{y-\hat{y}}$) outperform SeqCo(x-y) slightly. Given the fact that adding one more similarity loss increases around 30\% training time and the observations above, we recommend using a single similarity loss. 
We probably need to encourage the ``disagreement'' between them (we leave this for future work). As mentioned in Section \ref{sec:contrastive}, we can also use decoder based encoding function $f_\theta^{\text{D}}$ (see the SeqCo ($\lambda^D_{y-\hat{y}}$) and SeqCo ($\lambda_{y-\hat{y}}$) rows in Table \ref{tab:dev_cnndm}) and we obtain worse results. It may because influencing the decoding during contrastive training is too aggressive. Therefore, we only report results of contrastive models on single pair of text (i.e., SeqCo ($\lambda_{x-y}$), SeqCo ($\lambda_{x-\hat{y}}$) and SeqCo ($\lambda_{y-\hat{y}}$)) on NYT and XSum.
Again in Section \ref{sec:contrastive}, we propose to employ multi-head attention based similarity modeling (see Equation~(\ref{eq:multi-head}) and~(\ref{eq:sim})) rather than {\tt [CLS]} based method (see Equation~(\ref{eq:sim_cls})). It also shows attention based similarity, which takes associations across two sequences into account, is better (see SeqCo ($\lambda_{y-\hat{y}}$) and SeqCo ($\lambda_{y-\hat{y}}$) w/ {\tt [CLS]} rows in Table \ref{tab:dev_cnndm}).

\begin{table}[ht]
	\small
	\centering
	\begin{tabular}{l c c c}
		\hline
		Model &  R-1 & R-2 & R-L \\
		\hline 
		\multicolumn{4}{c}{Extractive}  \\
		\hline
		Lead3 & 39.58 & 20.11 & 35.78\\
		{\sc BertExt} & 46.66 & 26.35 & 42.62 \\
		\hline
		\multicolumn{4}{c}{Abstractive} \\
		\hline
% 		PTGen \cite{see-etal-2017-get} & 43.71 & 26.40 & - \\
        PTGen  & 43.71 & 26.40 & - \\
		{\sc BertSumExtAbs}  &49.02 & 31.02 & 45.55\\
		{\sc RoBERTa}-S2S & 45.92 & 29.48 & 42.73 \\
		STEP \cite{zou2020pre} & 50.03 & 32.12 & 46.25\\
		BART$\star$ \cite{lewis2020bart} & \textbf{53.20} & \textbf{35.04} & \textbf{49.23} \\
		\hline
		\multicolumn{4}{c}{Combination Methods} \\
		\hline
		GSum \cite{dou-etal-2021-gsum} & 54.27 & 35.37 & 47.63 \\
		\hline
		\multicolumn{4}{c}{Ours} \\
		\hline
		SeqCo ($\lambda_{x-y}$) &53.79 & 35.43  & 49.84 \\
		SeqCo ($\lambda_{x-\hat{y}}$) & \textbf{54.25}* & \textbf{35.82}* & \textbf{50.24}*  \\
		SeqCo ($\lambda_{y-\hat{y}}$)& 54.14 & 35.69 & 50.11 \\
		\hline
	\end{tabular}
	\caption{Results on the test split of NYT using limited-length recall based ROUGE. $\star$ means our own re-implementation. * means outperforms BART$\star$ significantly ($p<0.05$).}
	\label{tab:nyt}
\end{table}

Results on NYT are shown in Table \ref{tab:nyt} and the trend is similar. {\sc RoBERTa}-S2S is a transformer based Seq2Seq model with encoder initialized from RoBERTa \cite{liu2019roberta} and its results are reported in \cite{zou2020pre}. SeqCo ($\lambda_{x-\hat{y}}$) outperforms BART$\star$ by +1.0 ROUGE-1, +0.8 ROUGE-2 and +1.0 ROUGE-L and the differences between them are significant measured by the ROUGE script. SeqCo ($\lambda_{x-\hat{y}}$) obtains better results than all models in comparison. We again observe that using generated summaries in SeqCo are better than using gold summaries only.

% \subsection{Results on Xsum}
\begin{table}[ht]
	\small
	\centering
	\begin{tabular}{l c c c}
		\hline
		Model &  R-1 & R-2 & R-L \\
		\hline 
		\multicolumn{4}{c}{Extractive}  \\
		\hline
		Lead3 & 16.30& 1.60& 11.95\\
		{\sc MatchSum} & \textbf{24.86} & \textbf{4.66} & \textbf{18.41} \\
		\hline
		\multicolumn{4}{c}{Abstractive} \\
		\hline
% 		PTGen \cite{see-etal-2017-get} & 28.10 & 8.02 & 21.72\\
        PTGen & 28.10 & 8.02 & 21.72\\
		{\sc BertSumExtAbs} &38.81 & 16.50 & 31.27\\
		{\sc RoBERTa}-S2S & 43.54 & 20.49 & 35.75 \\
		STEP \cite{zou2020pre}  & 43.02 & 20.11 & 35.34\\
		PEGASUS (C4) & 45.20 & 22.06 & 36.99 \\
		PEGASUS (HugeNews) & \textbf{47.21} & \textbf{24.56} & \textbf{39.25} \\
		BART \cite{lewis2020bart} & 45.14 & 22.27 & 37.25 \\
		BART$\star$ \cite{lewis2020bart} & 45.35	& 22.01 & 36.76 \\
		\hline
		\multicolumn{4}{c}{{Combination Methods}} \\
		\hline
		GSum \cite{dou-etal-2021-gsum} & 45.40 & 21.89 & 36.67 \\
		simCLS \cite{liu-liu-2021-simcls} & \textbf{47.61} & \textbf{24.57} & \textbf{39.44} \\
		Refsum \cite{liu-etal-2021-refsum} & 47.45 & 24.55 & 39.41 \\ 
		\hline
		\multicolumn{4}{c}{Ours} \\
		\hline
		SeqCo ($\lambda_{x-y}$) & \textbf{45.65}* & \textbf{22.41}* & \textbf{37.04}* \\
		SeqCo ($\lambda_{x-\hat y}$) & 45.6 & 22.36 & 36.94 \\
		SeqCo ($\lambda_{y-\hat y}$) & 45.52 & 22.24 & 36.90 \\
		\hline
	\end{tabular}
	\caption{Results on the test split of XSum using full length F1 based ROUGE. $\star$ means our own re-implementation. * means outperforms BART$\star$ significantly ($p<0.05$).}
	\label{tab:xsum}
\end{table}

Table \ref{tab:xsum} summarizes our results on the XSum dataset. BART$\star$ (our reimplementation) are better at ROUGE-1, but worse at ROUGE-2 and \mbox{ROUGE-L} compared to BART. SeqCo ($\lambda_{x-y}$) outperforms BART$\star$ significantly measured with the ROUGE script. Results of SeqCo ($\lambda_{x-y}$) are better than all previously published models except for PEGASUS (HugeNews) and Refsum. It is not entirely surprising, because PEGASUS (HugeNews) is trained on 3,800 GB news data (the same genre as the XSum dataset), while PEGASUS(C4) is pre-trained on the C4 dataset consist of text from 350M Web pages (750GB) and performs  worse than PEGASUS (HugeNews). Refsum reranks outputs of PEGASUS (HugeNews). Note  that the pre-trained transformer (i.e., BART) in SeqCo is trained on only 160 GB data, which also contains data in other domains rather than news data. % {\bf why in XSum SeqCo ($\lambda_{x-y}$) works the best?}
\paragraph{Human Evaluation}

We do human evaluations on CNNDM, NYT and XSum with 100 documents each. We asked the participants to rank the outputs of different systems according to their faithfulness and the mean rank scores (lower is better) are shown in table~\ref{tab:human}. We employed (self-reported) native speakers to annotate our output summaries on Amazon Mechanical Turk. To further guarantee the annotation quality, we filter out the annotated assignments which were done less than two minutes (average time spent per assignment is 6 minutes). After the filtering process, we guarantee each document is annotated by three annotators. In CNNDM and NYT datasets, Seqco outperforms BART significantly. In XSum dataset, there are no significant differences among these systems. It may be because generated summaries in XSum are shorter, which are difficult for annotators to tell the differences. We calculate the ratios of agreement between annotators (i.e., ratio of all three annotators’ agreement and ratios of at least two annotators’ agreement) to measure the agreement for human evaluation. As shown in table~\ref{tab:aggrement}, there are around 30\% of summaries that all of 3 participants give the same annotations, and more than 90\% of summaries obtained the same annotations by at least 2 annotators. 
In addition, the Fleiss’ Kappa scores are 0.329 on CNNDM, 0.313 on NYT and 0.364 on XSum, which demonstrate a fair degree of agreement. 
We believe the agreement between annotators is reasonable.

\begin{table}[ht]
    \centering
    \begin{tabular}{c|c c c c}
    \hline
        systems & BART & $x-y$ & $x-\hat{y}$ & $y-\hat{y}$ \\
    \hline
        % CNNDM &  2.83 &  2.42 &  2.45 &  2.29\\
        CNNDM & 2.62 & 2.51 & 2.45* & 2.42* \\
        NYT   &  2.68 & 2.46*  & 2.39*  &  2.46* \\
        XSum  & 2.47 & 2.44  & 2.58  & 2.50 \\
    \hline
    \end{tabular}
    \caption{Human evaluation on faithfulness with mean rank (lower is better). We randomly sample 100 documents for each dataset and asked the participants to rank the outputs of different systems according to their faithfulness. * means this result is significantly different ($p<0.05$) from BART. }
    \label{tab:human}
\end{table}

\begin{table}[ht]
    \centering
    \begin{tabular}{c|ccc}
    \hline
    Datasets &  CNNDM & NYT & Xsum \\
    \hline
    3 agree  &   26.50\% & 31.00\% & 29.50\% \\
    $\ge2$ agree &  96.25\% & 95.75\% & 94.75\% \\
    \hline
    \end{tabular}
    \caption{The ratios of agreement between annotators.}
    \label{tab:aggrement}
\end{table}

\begin{table}[ht]
    \centering
    % \small
    \begin{tabular}{l c c c}
    \hline
        Model & 1-gram & 2-gram & 3-gram \\
        \hline
        \multicolumn{4}{c}{CNNDM} \\
        \hline
        Gold &  0.1360 & 0.4871 & 0.6908 \\
        BART & 0.0157 & 0.1140 &  0.2161 \\
        SeqCo & 0.0228 & 0.1524 & 0.2769 \\
        \hline
        \multicolumn{4}{c}{NYT} \\
        \hline
        Gold & 0.1064 & 0.4260 & 0.6189 \\
        BART & 0.0350 & 0.2231 & 0.3896 \\
        SeqCo & 0.0368 & 0.2284 & 0.3961 \\
        \hline 
        \multicolumn{4}{c}{XSum} \\
        \hline
        Gold & 0.3752 & 0.8328 & 0.9551 \\
        BART & 0.2821 & 0.7341 & 0.8924  \\
        SeqCo & 0.2929 & 0.7465 & 0.9015 \\
        \hline
    \end{tabular}
    \caption{Proportions of novel n-grams w.r.t. original documents in gold and model generated summaries on the validation sets of CNNDM, NYT and XSum.}
    \label{tab:novel_ngram}
\end{table}

\paragraph{Analysis} \emph{Different from CNNDM and NYT, why does using generated summaries in contrastive learning perform worse on XSum?} As shown in Table \ref{tab:novel_ngram}, it may because XSum is more abstractive (see the novel $n$gram statistics of Gold on the three datasets) and more difficult. 
% Summarization models are easier to learn copying from documents \cite{see-etal-2017-get,liu-lapata-2019-text}. 
As a result, the generated summaries are easier to have different meanings from their documents and gold summaries (at least in the early stage of training). Maybe that is the reason why the $x-\hat{y}$ and $y-\hat{y}$ objective is worse than the $x-y$ objective. CNNDM and NYT are less abstractive and the generated summaries could retain the main meanings more easily and are also more diverse (compared to gold summaries), which leads to the  $x-\hat{y}$ and $y-\hat{y}$ objectives work better.

We can also see from Table \ref{tab:novel_ngram} that SeqCo can either be more abstractive than BART or almost as abstractive as BART.
To choose the contrastive objective, our suggestion  is 1) for the datasets whose summaries are highly abstractive, choose the $x-y$ pair as the contrastive objective; 2) for less abstractive datasets (the case for most datasets), choose either $x-\hat{y}$ or $y-\hat{y}$ as the contrastive objective. As far as we observed, the performance of $x-\hat{y}$ and $y-\hat{y}$ are similar.

\paragraph{Ablation Study}
We list the ablation results on three datasets in the appendix A. We compared single similarity loss v.s. multiple similarity losses on the validation and test sets and observed the similar trends. We find best results are achieved by using a single similarity loss on all datasets except for the validation set of XSum, where SeqCo ($\lambda_{x-y}$ + $\lambda_{y-\hat{y}}$)  and SeqCo ($\lambda_{x-y}$ + $\lambda_{x-\hat{y}}$ + $\lambda_{y-\hat{y}}$) outperform SeqCo(x-y) slightly. Given the fact that adding one more similarity loss increases around 30\% training time and the observations above, we recommend using a single similarity loss.

\paragraph{Example Outputs}
Some example outputs of SeqCo and BART$\star$ are also listed in appendix B. In conclusion, BART sometimes miss some important points, while SeqCo can do better. 

\section{Conclusions}
In text summarization, a document, its gold summary and model generated summaries can be viewed as different views of the same meaning representation. We propose SeqCo, a sequence level contrastive learning model for text summarization, which intends to minimize distances between the document, its summary and its generated summaries during training. Experiments on three summarization datasets (CNNDM, NYT and XSum) show that SeqCo consistantly improves a strong Seq2Seq text generation model.
% \textcolor{red}{and SeqCo also achieves the state-of-the-art results on CNNDM and NYT.} 
In the future, we plan to extend SeqCo in the multi-lingual or cross-lingual text generation tasks. We observed in experiments that using multiple contrastive objectives did not improve the results. We are interested in developing methods for regularizing different contrastive objectives.

\bibliography{aaai22}

% \iffalse
\clearpage
\appendix
\section{Ablation Results}
We list the ablation results on the validation and test set for three datasets in table~\ref{tab:abla_cnndm}, \ref{tab:abla_nyt} and \ref{tab:abla_xsum}. We compared single similarity loss v.s. multiple similarity losses on the validation and test sets of the other two datasets and observed similar trends with CNNDM. We find best results are achieved by using a single similarity loss on all datasets except for the validation set of XSum, where SeqCo ($\lambda_{x-y}$ + $\lambda_{y-\hat{y}}$)  and SeqCo ($\lambda_{x-y}$ + $\lambda_{x-\hat{y}}$ + $\lambda_{y-\hat{y}}$) outperform SeqCo(x-y) slightly. Given the fact that adding one more similarity loss increases around 30\% training time and the observations above, we recommend using a single similarity loss.

\section{Examples}
We list some examples of generated summaries and gold summaries in table~\ref{tab:example1} and \ref{tab:example2} on the test set of CNNDM, where we can compare the outputs of BART and SeqCo. In conclusion, BART sometimes miss some important points, while SeqCo can do better. 

In the document of Table~\ref{tab:example1}, an important point is that Anne Frank and her older sister died earlier than previously believed. The output of BART doesn't mention this directly but describes two dates of their death, which is lack of the main idea and confusing. SeqCo points out this emphasis in the first sentence and then further explains, which is quite consistent with the meaning expressed by the gold summary.

For the document in Table~\ref{tab:example2}, the most important thing is that Schuller died. BART focuses on what did do and doesn't mention the death, while the gold summary and SeqCo both describe his death in the first sentence, and then list some famous deeds in his lifetime. Death is more important than the deeds in this document.

\begin{table*}[hb]
% 	\small
	\centering
	\begin{tabular}{l|c c c  c c c}
		\hline
		& \multicolumn{3}{c}{Validation set} & \multicolumn{3}{c}{Test set} \\
		 &  R-1 & R-2 & R-L &  R-1 & R-2 & R-L \\
		\hline
		BART$\star$ & 45.24 & 22.10 & 42.01 &   44.10 & 21.31 & 40.91 \\
		SeqCo ($\lambda_{x-y}$) & 45.60 & 22.30 & 42.36 &  44.66 & 21.57 & 41.38 \\
	    SeqCo ($\lambda_{x-\hat{y}}$) & 45.80 & 22.39 & 42.57 & 44.94 & \textbf{21.82} & 41.68 \\
		% BART + $\lambda_b(<s>)$ &  \\
		SeqCo ($\lambda_{y-\hat{y}}$) & \textbf{45.88} & \textbf{22.46} & \textbf{42.66} & \textbf{45.02} & 21.80 & \textbf{41.75} \\
		SeqCo ($\lambda_{y-\hat{y}}$) w/ {\tt [CLS]} & 45.72 & 22.42 & 42.48 & 44.81 & 21.70 & 41.56  \\
		SeqCo ($\lambda_{x-y}$ + $\lambda_{x-\hat{y}}$) & 45.68 & 22.38 & 42.45 & 44.95 & {21.81} & 41.68 \\
		SeqCo ($\lambda_{x-y}$ + $\lambda_{y-\hat{y}}$)& 45.62 & 22.29 & 42.37 & 44.86 & 21.77 & 41.58 \\
		SeqCo ($\lambda_{x-\hat{y}}$ + $\lambda_{y-\hat{y}}$) & 45.72 & 22.35 & 42.45 & 44.85 & 21.72 & 41.58 \\
	    SeqCo ($\lambda_{x-y}$ + $\lambda_{x-\hat{y}}$ + $\lambda_{y-\hat{y}}$) & 45.72 & 22.38 & 42.46 & 44.73 & 21.67 & 41.45 \\
		SeqCo ($\lambda^D_{y-\hat{y}}$) & 45.74 & 22.39 & 41.55 & 44.86 & 21.66 & 41.55\\
		\hline
	\end{tabular}
	\caption{Results on the test split of CNNDM using full length F1 based ROUGE-1 (R-1), ROUGE-2 (R-2) and
		ROUGE-L (R-L). SeqCo ($\lambda_{x-y}$), SeqCo ($\lambda_{x-\hat{y}}$) and SeqCo ($\lambda_{y-\hat{y}}$) stand for contrastive learning between document and gold summary, document and generated summary as well as gold and generated summary. $\star$ means our own re-implementation.}
	\label{tab:abla_cnndm}
\end{table*}

\begin{table*}[hb]
% 	\small
	\centering
	\begin{tabular}{l|c c c  c c c}
		\hline
		& \multicolumn{3}{c}{Validation set} & \multicolumn{3}{c}{Test set} \\
		 &  R-1 & R-2 & R-L &  R-1 & R-2 & R-L \\
		\hline
		BART$\star$ & 50.75 & 31.71 & 46.32 & 53.20 &  35.04 & 49.23 \\
		SeqCo ($\lambda_{x-y}$) &50.85 & 31.63  & 46.38 &  53.79 & 35.43 & 49.84  \\
		SeqCo ($\lambda_{x-\hat y}$) & \textbf{51.27} & \textbf{31.99} & \textbf{46.74} & \textbf{54.25} & \textbf{35.82} & \textbf{50.24}  \\
		SeqCo ($\lambda_{x-\hat y}$) w/ {\tt [CLS]} &50.79 & 31.61 & 46.33 &  53.70 & 35.33 & 49.78  \\
		SeqCo ($\lambda_{y-\hat y}$) & 51.38 & 32.01 & 46.87 & 54.14 & 35.69 & 50.11 \\
		SeqCo ($\lambda_{x-y} + \lambda_{x-\hat y}$) &50.88 & 31.61 & 46.39 & 53.82 & 35.43 & 49.85  \\
		SeqCo ($\lambda_{x-y} + \lambda_{y-\hat y}$)  & 50.97 & 31.70 & 46.43 & 53.79 & 35.34 & 49.79 \\
		SeqCo ($\lambda_{x-\hat y} + \lambda_{y-\hat y}$) & 51.08 & 31.74 & 46.58 & 53.90 & 35.36 & 49.91  \\
		SeqCo ($\lambda_{x-y} + \lambda_{x-\hat y} + \lambda_{y + \hat y}$ ) & 51.21  & 31.91 & 46.65 & 53.95 & 35.54 & 49.97 \\
		\hline
	\end{tabular}
	\caption{Results on the validation and test split of NYT using limited-length recall based ROUGE. $\star$ means our own re-implementation.}
	\label{tab:abla_nyt}
\end{table*}

\begin{table*}[hb]
	\centering
	\begin{tabular}{l|c c c  c c c}
		\hline
		& \multicolumn{3}{c}{Validation set} & \multicolumn{3}{c}{Test set} \\
		 &  R-1 & R-2 & R-L &  R-1 & R-2 & R-L \\
		\hline
		BART$\star$ & 45.38 & 22.13 & 36.80 & 45.35 & 22.01 & 36.76  \\
		SeqCo ($\lambda_{x-y}$) & \textbf{46.66} & 22.42 & 37.14 & \textbf{45.65} & \textbf{22.41} & \textbf{37.04}  \\
		SeqCo ($\lambda_{x-\hat y}$) & 45.59 & 22.39 & 37.05 & 45.60 & 22.36 & 36.94   \\
		SeqCo ($\lambda_{x-\hat y}$) w/ {\tt [CLS]} &  45.41 & 22.25 & 36.97 & 45.30 & 22.15 & 36.78  \\ 
		SeqCo ($\lambda_{y-\hat y}$) & 45.59 & 22.39 & 37.08 & 45.52 & 22.24 & 36.90  \\
		SeqCo ($\lambda_{x-y} + \lambda_{x-\hat y}$) & 45.60 & 22.41 & 37.11 & 45.28 & 22.05 & 36.67   \\
		SeqCo ($\lambda_{x-y} + \lambda_{y-\hat y}$)  & 45.67 & 22.46 & \textbf{37.19} & 45.58 & 22.32 & 36.97  \\
		SeqCo ($\lambda_{x-\hat y} + \lambda_{y-\hat y}$) & 45.67 & 22.37 & 37.01 & 45.50 & 22.25 & 36.87 \\
		SeqCo ($\lambda_{x-y} + \lambda_{x-\hat y} + \lambda_{y + \hat y}$ ) & 45.77 & \textbf{22.52} & 37.16 & 45.51 & 22.22 & 36.87  \\
		\hline
	\end{tabular}
	\caption{Results on the validation and test split of XSum using full length F1 based ROUGE. $\star$ means our own re-implementation.}
	\label{tab:abla_xsum}
\end{table*}

\begin{table*}
    \centering
    \small
    \begin{tabular}{m{2cm}<{\centering}|  m{12cm}}
        \hline
        Article & Seventy years ago, Anne Frank died of typhus in a Nazi concentration camp at the age of 15. Just two weeks after her supposed death on March 31, 1945, the Bergen-Belsen concentration camp where she had been imprisoned was liberated -- timing that showed how close the Jewish diarist had been to surviving the Holocaust. But new research released by the Anne Frank House shows that Anne and her older sister, Margot Frank, died at least a month earlier than previously thought. Researchers re-examined archives of the Red Cross, the International Training Service and the Bergen-Belsen Memorial, along with testimonies of survivors. They concluded that Anne and Margot probably did not survive to March 1945 -- contradicting the date of death which had previously been determined by Dutch authorities. In 1944, Anne and seven others hiding in the Amsterdam secret annex were arrested and sent to the  Auschwitz-Birkenau concentration camp. Anne Frank's final entry . That same year, Anne and Margot were separated from their mother and sent away to work as slave labor at the Bergen-Belsen camp in Germany. Days at the camp were filled with terror and dread, witnesses said. The sisters stayed in a section of the overcrowded camp with no lighting, little water and no latrine. They slept on lice-ridden straw and violent storms shredded the tents, according to the researchers. Like the other prisoners, the sisters endured long hours at roll call. Her classmate, Nannette Blitz, recalled seeing Anne there in December 1944: ``She was no more than a skeleton by then. She was wrapped in a blanket; she couldn't bear to wear her clothes anymore because they were crawling with lice." Listen to Anne Frank's friends describe her concentration camp experience . As the Russians advanced further, the Bergen-Belsen concentration camp became even more crowded, bringing more disease. A deadly typhus outbreak caused thousands to die each day. Typhus is an infectious disease caused by lice that breaks out in places with poor hygiene. The disease causes high fever, chills and skin eruptions. ``Because of the lice infesting the bedstraw and her clothes, Anne was exposed to the main carrier of epidemic typhus for an extended period," museum researchers wrote. They concluded that it's unlikely the sisters survived until March, because witnesses at the camp said the sisters both had symptoms before February 7. ``Most deaths caused by typhus occur around twelve days after the first symptoms appear," wrote  authors Erika Prins and Gertjan Broek. The exact dates of death for Anne and Margot remain unclear. Margot died before Anne. ``Anne never gave up hope," said Blitz, her friend. ``She was absolutely convinced she would survive." Her diary endures as one of the world's most popular books. Read more about Anne Frank's cousin, a keeper of her legacy . \\
        \hline
        Gold Summary & Museum : Anne Frank died earlier than previously believed . Researchers re-examined archives and testimonies of survivors . Anne and older sister Margot Frank are believed to have died in February 1945 . \\
        \hline
        BART & Anne Frank died of typhus in a Nazi concentration camp at the age of 15 in 1945 . The date of death had previously been determined by Dutch authorities . Researchers re-examined archives of the Red Cross, the International Training Service and the Bergen-Belsen Memorial . They concluded that Anne and Margot probably did not survive to March 1945 . \\
        \hline
        SeqCo & New research shows Anne Frank and her older sister, Margot Frank, died at least a month earlier than previously thought . Researchers re-examined archives of the Red Cross, the International Training Service and the Bergen-Belsen Memorial . They concluded that Anne and Margot probably did not survive to March 1945 -- contradicting the date of death which had been determined . \\
        \hline

    \end{tabular}
    \caption{An example document and its gold summary sampled from the test splitting of CNNDM along with the outputs of BART and SeqCo.}
    \label{tab:example1}
\end{table*}

\begin{table*}[ht]
    \centering
    \small
    \begin{tabular}{m{2cm}<{\centering}|  m{12cm}}
        \hline
        Article & The Rev. Robert H. Schuller, California televangelist and founder of the television ministry ``Hour of Power," died Thursday, according to his family. He was 88 years old. Schuller, also the founder of Crystal Cathedral megachurch, had been diagnosed with esophageal cancer in August 2013, a release from ``Hour of Power" said. ``My father-in-law passed away peacefully early this morning. He was a great Dad and a great man of God," said Schuller's daughter-in-law, Donna Schuller, in a Twitter message. Schuller's life followed an almost Shakespearean arc. He was born in a Iowa farmhouse without running water and longed to preach from his earliest days. In his autobiography, ``Prayer: My Soul's Adventure with God," he described standing alone by a river and picturing himself delivering sermons to a rapt congregation. After attending a Hope College and Western Theological Seminary in Michigan, he met his wife of more than 60 years, Arvella, while preaching at her church (she was the organist). With their young family in tow, the Schullers caravanned west to California, where he rented a drive-in theater and preached from the roof of the snack bar. It was beneath the dignity of Christian ministry, some local pastors huffed. The ``passion pits" where teenagers necked was no place for the gospel. Schuller was undeterred, and he quickly outgrew the drive-in. He called the explosive growth of his tiny congregation a ``miracle," though his many mainstream critics had other names for it. His confident, breezy version of Christianity -- too breezy, by some estimations -- drew hordes of seekers and lapsed Christians who were put off by the hellfire fulminations of many post-War American preachers. Schuller sold a softer, gentler message, which borrowed heavily, he acknowledged, from the father of the feel-good gospel, Norman Vincent Peale. He preached not to convert or condemn people, but to encourage them, a sentiment he called ``possibility thinking." People loved it. ``Evangelicalism at its best wants to be innovative and reach people," said Timothy Larsen, a professor of Christian thought at Wheaton College in Illinois. ``And Schuller was a master at that." ``What he got right is that the gospel is good news," Larsen continued. ``And he preached an uplifting message about personal transformation and uplift and hope." Some of Schuller's favored phrases, though, struck others as cornpone Christianity. ``Turn your hurt into a halo?" said Randall Balmer, a professor of American religious history at Dartmouth College, citing one such phrase. ``That's pretty weak tea." Still, Balmer gives Schuller some credit. ``It may be bad theology, but it's brilliant marketing." In 1970, Schuller began broadcasting ``Hour of Power," believed to be one of the first, if not the very first, Sunday service to be shown regularly on television. With his genial smile, priestly robes and gray hair, he looked and talked like a guy who wanted nothing more than to see his flock succeed. The show, which ran for decades, reached millions, making Schuller a televangelist before the term became tarnished by the sins of his many successors. Schuller's crowning achievement, at least architecturally, still stands in Orange County, California, though it is now owned by the Roman Catholic Church. The Crystal Cathedral, a great gleaming edifice with 10,000 glass panels, gave worshipers a look at the  clouds that house the heavens, while Schuller preached in the pulpit below. The message was clear to many: The road to the former ran through the latter. During the 1980s and 1990s, Schuller's star continued to rise, with presidents stopping by the Crystal Cathedral -- often during campaigns, it should be said -- and future megachurch pastors like Rick Warren and Bill Hybels seeking his advice. As Schuller aged, though, his family was beset by a succession scandal straight from the pages of ``King Lear."  He tried to install his only son, Bobby Jr., as pastor of Crystal Cathedral. But the preaching styles of father and son were too different for the congregation -- measured at times at 10,000 strong -- to countenance. Bobby Schuller Jr. left ``Hour of Power" and the pulpit at Crystal Cathedral after a short time. As the family searched for a new successor and tussled over finances, viewers and donations to the church and its television show dropped precipitously. Crystal Cathedral Ministries filed for bankruptcy in 2010, citing debts of more than \$43 million, according to The Associated Press. Schuller's empire, which once soared as high as his glassy cathedral, had fallen to dust. Eventually, Schuller's grandson, also named Bobby, took over ``Hour of Power," though at a different church. In a statement on Thursday, the younger Schuller recalled standing atop Crystal Cathedral's 12-story Tower of Hope with his grandfather as they surveyed the surrounding landscape. ``You could see the whole world from there," he said. People we've lost in 2015 . CNN's Stella Chan reported from Los Angeles.\\
        \hline
        Gold Summary & The Rev. Robert Schuller , 88 , had been diagnosed with esophageal cancer in 2013 . His TV show , `` Hour of Power , '' was enormously popular in the 1970s and 1980s . \\
        \hline
        BART & The Rev. Robert H. Schuller had been diagnosed with esophageal cancer in August 2013 . He was the founder of the television ministry ``Hour of Power'' and the Crystal Cathedral megachurch . He sold a softer, gentler message, which borrowed heavily from the father of the feel-good gospel . \\
        \hline
        SeqCo & The Rev. Robert H. Schuller died Thursday at 88, his family says . He was the founder of the television ministry ``Hour of Powe'' and the Crystal Cathedral megachurch . He had been diagnosed with esophageal cancer in August 2013, the ministry says .\\
        \hline

    \end{tabular}
    \caption{An example document and its gold summary sampled from the test splitting of CNNDM along with the outputs of BART and SeqCo.}
    \label{tab:example2}
\end{table*}
% \fi

\end{document}